\documentclass[letterpaper]{article} 
\usepackage{aaai24}  
\usepackage{times}  
\usepackage{helvet}  
\usepackage{courier}  
\usepackage[hyphens]{url}  
\usepackage{graphicx} 
\usepackage{natbib}  
\usepackage{caption} 
\usepackage{times}
\usepackage{latexsym}

\usepackage{amsmath}
\usepackage{float}
\usepackage[caption = false]{subfig}
\usepackage{algorithm}
\usepackage{algorithmic}
\usepackage{amsmath,amssymb}
\usepackage{dsfont}
\usepackage{multirow}

\usepackage[T1]{fontenc}

\usepackage{fancyhdr}

\usepackage{newfloat}
\usepackage{listings}
\DeclareCaptionStyle{ruled}{labelfont=normalfont,labelsep=colon,strut=off} 
\floatstyle{ruled}
\newfloat{listing}{tb}{lst}{}
\floatname{listing}{Listing}
\title{Reinforcement Learning for Optimizing RAG for Domain Chatbots}
\author{
    Mandar Kulkarni,
    Praveen Tangarajan,
    Kyung Kim,
    Anusua Trivedi
}
\affiliations{
    Flipkart Data Science\\

    Seattle, Washington, USA\\
}

\usepackage{bibentry}

\begin{document}

\maketitle

\thispagestyle{fancy}
\fancyhf{}
\rhead{Accepted at AAAI 2024 Workshop on Synergy of Reinforcement Learning and Large Language Models}

\begin{abstract}

With the advent of Large Language Models (LLM), conversational assistants have become prevalent for domain use cases.
LLMs acquire the ability to contextual question answering through extensive training, and Retrieval Augmented Generation (RAG) further enables the bot to answer domain-specific questions. This paper describes a RAG-based approach for building a chatbot that answers user's queries using Frequently Asked Questions (FAQ) data. We train an in-house retrieval embedding model using infoNCE loss, and experimental results demonstrate that the in-house model works significantly better than the well-known general-purpose public embedding model, both in terms of retrieval accuracy and Out-of-Domain (OOD) query detection. As an LLM, we use an open API-based paid ChatGPT (gpt-35-turbo-16k-0613) model.
We noticed that a previously retrieved-context could be used to generate an answer for specific patterns/sequences of queries (e.g., follow-up queries). Hence, there is a scope to optimize the number of LLM tokens and cost. Assuming a fixed retrieval model and an LLM, we optimize the number of LLM tokens using Reinforcement Learning (RL). Specifically, we propose a policy-based model external to the RAG, which interacts with the RAG pipeline through policy actions and updates the policy to optimize the cost. The policy model can perform two actions: to fetch FAQ context or skip retrieval. We use the open API-based GPT-4 as the evaluation model, which rates the quality of the bot answer. Using an appropriate reward shaping, GPT-4 ratings are converted to a numeric reward. We then train a policy model using policy gradient on multiple training chat sessions. As a policy model, we experimented with a public gpt-2 model and an in-house BERT model. With the proposed RL-based optimization combined with similarity threshold, we are able to achieve significant (\textasciitilde{31\%}) cost savings while getting a slightly improved accuracy. Though we demonstrate results for the FAQ chatbot, the proposed RL approach is generic and can be experimented with any existing RAG pipeline.

\end{abstract}

\section{Introduction}


With the advent of Large Language Models (LLM), we observe an increased use of conversational assistants even for the domain use cases. 
Trained on a large web-scale text corpus with approaches such as instruction tuning and Reinforcement Learning with Human Feedback (RLHF), LLMs have become good at contextual question-answering tasks, i.e., given a relevant text as a context, LLMs can generate answers to questions using that information.
Retrieval Augmented Generation (RAG) is one of the key techniques used to build chatbots for answering questions on domain data. RAG consists of two components: a retrieval model and an answer generation model based on LLM. The retrieval model fetches context relevant to the user's query. 
The query and the retrieved context are then inputted to the LLM with the appropriate prompt to generate the answer.
Though intuitive to understand, RAG-based chatbots have a few challenges from a practical standpoint.

\begin{enumerate}
\item If a retrieval model fails to fetch relevant context, the generated answer would be incorrect and uninformative. The retrieval is more challenging for non-English queries, e.g., code-mix Hinglish. 

\item For paid API-based LLMs (e.g., ChatGPT), the cost per call is calculated based on the number of input and output tokens. A large number of tokens passed in a context leads to a higher cost per API call. With a high volume of user queries, the cost can become significant.


\item To enable multi-turn conversations with RAG, a conversation history needs to be maintained and passed to the LLM with every query. It is known that a larger input token size leads to a drop in accuracy or hallucinations as LLMs have an additional task of choosing relevant information from a large context \cite{liu2023lost}. 

\end{enumerate}

In this paper, we first describe a RAG-based approach for building a chatbot that answers user's queries using Frequently Asked Questions (FAQ) data. We have a domain FAQ dataset consisting of 72 FAQs regarding the credit card application process. The FAQ dataset is prepared to answer user queries regarding general card information pre- and post-application queries. We train an in-house retrieval (embedding) model using info Noise Contrastive Estimation (infoNCE) loss \cite{oord2019representation} with the English and Hinglish paraphrase queries created using ChatGPT. The embedding model is trained to maximize query-question and query-QnA similarity. Experimental results show that the in-house model performs significantly better than a public pre-trained embedding model regarding retrieval accuracy and Out-of-Domain (OOD) query detection. OOD queries are questions unrelated to the domain data, e.g., how is the weather today? For every user query, we retrieve top-k FAQs (question + answer) as context and input them to the LLM to generate the answer. We maintain the previous two queries, answers, and FAQ context as history to enable the multi-turn conversation. We use an open paid API-based ChatGPT (gpt-35-turbo-16k-0613) for all our experiments as an LLM. GPT-4 is used to rate the bot answer quality (as Good/Bad) for the end-to-end RAG pipeline.

We propose an RL-based approach to optimize the number of tokens passed to the LLM. We noticed that for certain patterns/sequences of queries, we can get a good answer from the bot even without fetching the FAQ context. Examples of such scenarios can be: 1. for a follow-up query; FAQ context need not be retrieved if it has already been fetched for the previous query; 2. for the sequence of queries referring to the same FAQ, a context can be fetched only once at the start; 3. for OOD queries, the LLM prompt itself can guide the bot to generate the answer. Using this insight, we propose a policy gradient-based approach to optimize the number of LLM tokens and, hence, the cost. The input to the policy model is the State, which comprises previous queries, previous policy actions, and the current query. The policy model can take two actions: [FETCH] or [NO\textunderscore FETCH]. A usual RAG pipeline would be executed when a policy network takes a [FETCH] action. When a policy network chooses a [NO\textunderscore FETCH] action, FAQ context is not retrieved. A query and context (empty in the case of [NO\textunderscore FETCH] action) are inputted to the LLM. We use GPT-4 as the reward model and convert the quality rating (Good/Bad) to the numeric reward using appropriate reward shaping. If the LLM generates a good answer (as rated by GPT-4) even without fetching the context, we give it a high positive reward to promote such actions in the future. If not fetching a FAQ context leads to a wrong answer, we provide a negative reward (e.g., this can happen if the policy model chooses a  [NO\textunderscore FETCH] action for a domain query without any previous relevant context). For the training chat sessions, for each State, a policy model samples an action according to the current probability distribution over the actions. We then generate (State, Action, Reward) trajectories by sampling multiple times from the current policy. A policy model is then updated using a policy gradient with cumulative reward. As a policy model, we experimented with in-house BERT and public gpt-2 models. Figure \ref{fig:pg} shows the architecture of the proposed policy-based approach. Experimental results demonstrate that the policy model provides token saving by fetching the FAQ context only when it is required. When combined with a simple similarity threshold-based optimization, we are able to achieve token savings of \textasciitilde{31\%} on the test chat session with 91 queries while achieving slightly improved accuracy (evaluated through manual labeling) than the usual RAG pipeline. This underlines the effectiveness of the proposed approach. 

\section{Related works}

With recent advancements in Generative AI and LLMs, the Retrieval-Augmented Generation (RAG) \cite{lewis2021retrievalaugmented} approach has emerged as the preferred strategy for contextual question answering. 
RAG pipeline consists of a retrieval model followed by an LLM to generate the answer. Different approaches have been experimented with to improve components of the RAG in terms of accuracy and minimize hallucinations in answer generation.
Khatry et al. \cite{khatry2023augmented} proposed a low-rank residual adaptation approach with the pre-trained embedding model to improve the retrieval model. It was shown to lead to improved task-specific retrieval as compared to a general-purpose embeddings-based baseline.
Instead of using an interleaved retrieval and generation, Shao et al. \cite{shao2023enhancing} proposed an iterative retrieval and generation approach where the current model output acts as an informative context for retrieving more relevant knowledge which in turn helps generate a better output in the next iteration. Li et al. \cite{li2022survey} extensively surveys recent RAG-based approaches.


RL has been experimented with to improve RAG.
Bacciu et al. \cite{rraml} propose an RL-based approach to train an efficient retriever model to search for relevant information in an arbitrarily large database. Once this set of relevant data has been retrieved, it is forwarded to the API-based LLM to generate the answer. In particular, the authors show that RL helps reduce hallucinations by minimizing the number of damaging documents returned by the retriever. Self-RAG \cite{asai2023selfrag} trains a single LLM that adaptively retrieves passages and generates and reflects on retrieved passages and their generations using reflection tokens. In our work, we assume we do not have access to the gradients of the retrieval model and LLM. We only train a policy model that resides external to RAG.

GPT-4 is observed to provide human-level accuracy for automated evaluation tasks. Hack at al. \cite{hackl2023gpt4} investigated the consistency of feedback ratings generated by GPT-4 across multiple iterations, time spans, and stylistic variations. The study indicated that GPT-4 can generate consistent ratings across repetitions with a clear prompt.
Liu et al. \cite{liu2023geval} evaluated GPT-4 with two generation tasks: text summarization and dialogue generation. It was shown that GPT-4 provided state-of-the-art results on automated evaluation, which highly correlates with human evaluation. In our work, we use GPT-4 for automated evaluation of bot responses without ground truth answers.

\section{RAG for FAQ chatbots} 
\pagestyle{plain}


In this section, we describe details of the datasets and embedding model training for RAG for FAQ chatbots. 

\subsection{Dataset}

We have a FAQ dataset consisting of 72 FAQs regarding the credit card application process. The FAQ dataset is prepared to answer user queries regarding general card information (e.g., annual fee, benefits), pre-application issues faced on the platform (e.g., server issue, content not available, not getting One Time Password (OTP)), and post-application services (e.g., increasing credit limit, purchase EMI options). The FAQs list is designed based on the user survey conducted on the platform, collecting information about current pain points faced by users about the application.  



\subsection{Training in-house embedding model}
\label{sec:trin}

To use RAG for FAQ chatbots, we first need to retrieve the top k most relevant FAQs to answer the given query. To fetch top-k FAQs, an embedding model is used to encode queries and FAQs. We concatenate a question and an answer within a FAQ and use this to get a vector for that FAQ. A cosine similarity measure is then used to rank the FAQs based on their relevance. We experimented with popular general-purpose public pre-trained embedding models and models trained on in-domain data. We trained an in-house embedding model with infoNCE loss \cite{oord2019representation} using a well-known embedding model (e5-base-v2) weights as initialization. InfoNCE has proved an effective loss for contrastive learning, specifically under self-supervised setting \cite{goyal2021selfsupervised}. 
We finetune a publicly available embedding model \textit{e5-base-v2} \cite{wang2022text} using in-domain data with infoNCE loss. 

We create a dataset for training the embedding using ChatGPT and small manual tagging. We generate multiple English queries per FAQ as question paraphrases. Due to the large non-English-speaking population in India, we observe many Hinglish queries on the platform. We also generate a few Hinglish paraphrase queries per FAQ to support code-mix Hinglish queries. Next, we add a small set of manually tagged corpus where we add queries that can be answered based on the FAQ answers' content. We use \textasciitilde{3.5k} queries for training, \textasciitilde{1k} queries for validation, and 1014 queries for testing.

We use the following loss function to train the model.

\begin{eqnarray}\label{eq:so1}
l_{i,j} = \frac{\exp( \text{sim}(z_i,z_j)/ \tau ) }{\sum_{k=1}^{2B} \mathds{1}_{[k \neq i]} \exp( \text{sim}(z_i,z_k)/ \tau )}
\end{eqnarray}
Here, $z_i$ and $z_j$ indicate a positive pair (i.e., question paraphrases for the same FAQ), $B$ indicates batch size, and $\tau$ indicates temperature. For all our experiments, we set $N$ to 8 and $\tau$ to 0.1 and use cosine similarity for $\text{sim}$.


As seen from Eq. \ref{eq:so1}, infoNCE loss is specifically suitable for training with query-FAQ mapped data because it only needs positive pairs and treats remaining samples as in-batch negatives. We finetune the embedding model with two objectives: maximize query-QnA similarity and query-question similarity.  

Table \ref{tab:ae} shows the top-1 and top-3 accuracy comparison results for English and Hinglish queries with a general-purpose public model (e5-base-v2), the models finetuned with triplet loss and infoNCE loss. The ranking results are computed based on query-QnA cosine similarity. It can be seen that the in-house model trained with infoNCE works significantly better than the public pre-trained model. 

\begin{table}[h]
    \centering
    \captionsetup{justification=centering, margin=5mm}
    \begin{tabular}{|l|l|l|l|l|}
   \hline 
   \multicolumn{1}{|p{1.5cm}|}{\textbf{Model}} &
      \multicolumn{2}{c|}{\textbf{English}} &
      \multicolumn{2}{c|}{\textbf{Hinglish}}\\
      \hline
     & top-1 & top-3 & top-1 & top-3  \\
    \hline
    e5-base-v2 & 0.82 & 0.91 & 0.71 & 0.87  \\
    \hline
    triplet-loss & 0.90 & 0.93 & 0.84 & 0.89  \\
    \hline
    infoNCE & 0.97 & \textbf{0.98} & 0.94 & \textbf{0.95}  \\
    \hline
        
\end{tabular}  
    \vspace{2mm}
    \caption{Retrieval accuracy comparison of public vs in-house embedding models}
    \label{tab:ae}
\end{table}



Next, we compare the detection performance of public and in-house models on in-domain vs out-of-domain (OOD) samples.
We created a dataset of 30 in-domain and OOD queries each. In-domain queries contain English and Hinglish queries, while OOD queries include greeting, acknowledgment, and general non-domain-related questions. We calculated mean scores for positive and negative queries with the public and finetuned model. Table \ref{tab:ms} shows the average of top-1 matching scores for the two models. Even though the in-house model is not specifically trained on OOD queries, it performs much better in OOD query identification. A general-purpose pre-trained model assigns less discrimination scores for in-domain and OOD queries. Therefore, with a public embedding model, it's not possible to use a similarity threshold. As shown later, such a threshold can help token optimization for FAQ chatbots.





\begin{table}[h]
    \centering
    \captionsetup{justification=centering, margin=5mm}
    \begin{tabular}{|c|c|c|}
        
        \hline
        \textbf{Model} & \textbf{In-domain} & \textbf{OOD} \\
        \hline
        e5-base-v2 & 0.84 & 0.82 \\
        \hline
        finetuned with InfoNCE & 0.85 & \textbf{0.56} \\
        \hline
        
        \end{tabular}
    \vspace{2mm}
    \caption{Comparison of top-1 cosine similarity scores for in-domain vs OOD queries}
    \label{tab:ms}
\end{table}

\begin{figure*} 
\centering

\begin{tabular}{c}

\includegraphics[width=\linewidth, height = 80pt]{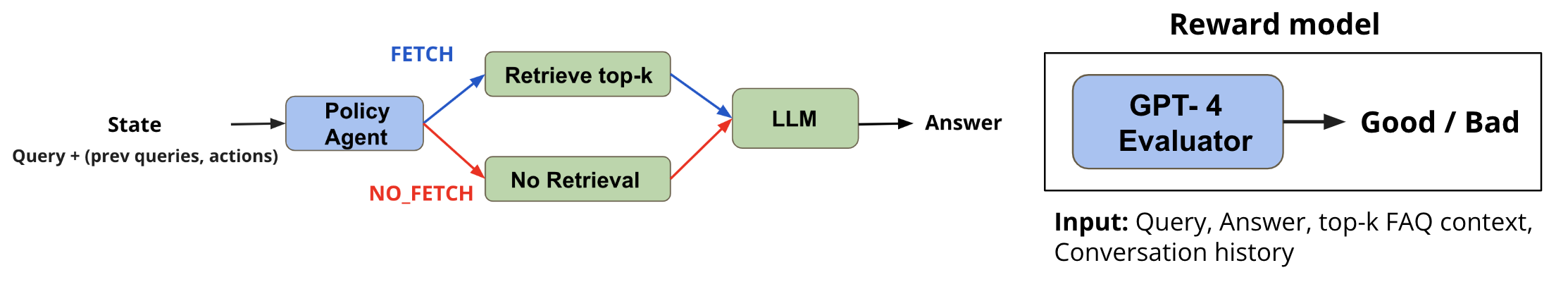}

\end{tabular}
\caption{Proposed policy agent based architecture for optimizing RAG for domain chatbots.}
\label{fig:pg}
\end{figure*}

\subsection{Prompt for ChatGPT}

Since this is a finance-related bot, we attempted to design a prompt that minimizes hallucinations with ChatGPT. Apart from adding a relevant prompt for contextual QnA, we add in-context examples to get the results in the required JSON format. Adding a post prompt was observed to help with reducing hallucinations. A post prompt is a small text instruction passed to ChatGPT along with every query. In our case, the post prompt is 'Instruction: Answer only using the information available.'. To avoid providing irrelevant information for OOD queries, ChatGPT is prompted to provide a "do not know" answer.



For all our experiments, we set k to 3. For a ChatGPT call, we maintain the last two conversations as a history. The conversation history includes previous queries, generated bot answers, and faq context. We need to maintain FAQ context in conversation history to answer follow-up questions.
E.g. for the sequence of two queries ['is there annual fee',  'can you reduce it?'], a FAQ context fetched for the second query would be irrelevant, and ChatGPT needs to rely on the FAQ context fetched for the first query. 




\subsection{Evaluation}

With the lack of ground truth answers with a generative model such as ChatGPT, off-the-shelf evaluation metrics such as BLEU or ROUGE cannot be used. To evaluate the quality of answers generated from the bot, we use the GPT-4-based approach. GPT-4 has shown promising results for automated quality evaluation compared to human evaluation  \cite{hackl2023gpt4} \cite{liu2023geval}. 
We provide GPT-4 with the current query, current FAQ context, answer generated from the bot, and conversation history (which includes previous query, answers, and FAQ context). We do not provide GPT-4 with the ideal or ground truth response. Instead, we prompt GPT-4 to evaluate the quality of the generated answer based on the information available in the current FAQ context and conversation history. GPT-4 is instructed to evaluate the output based on whether it exactly matches the information provided in the FAQ context.
Any additions or deviations from the available information are termed as factual failures and are to be rated bad. GPT-4 is asked to provide a final rating as good or bad and the reason behind the rating. For the OOD queries, the bot is supposed to provide a "do not know" answer, and GPT-4 is instructed to rate such answers as good. Table \ref{tab:gpt2} and \ref{tab:gpt1} show query, bot response, and GPT-4 evaluation examples.

\begin{table}[!h]
    \centering
    \captionsetup{justification=centering, margin=5mm}
    \begin{tabular}{|c|p{4.5cm}|}
        
        \hline
        \textbf{query} & is there annual fee for card \\
        \hline
        \textbf{bot response} & Yes, there is an annual fee of INR 500 for the credit card. However, this fee will be waived off if your annual spends exceed INR 3.5 Lakhs. \\
        \hline
        \textbf{gpt4 eval} & Good \\
        \hline
        \textbf{gpt4 reason} & The answer is directly addressing the query and providing the relevant information from the FAQ context. The answer matches with the information in the FAQ context, stating that there is an annual fee of INR 500 for the the credit card, which will be waived off if the annual spends exceed INR 3.5 Lakhs. \\
        \hline
        
        \end{tabular}
    \vspace{2mm}
    \caption{GPT-4 evaluation for an in-domain query}
    \label{tab:gpt2}
\end{table}

\begin{table}[!h]
    \centering
    \captionsetup{justification=centering, margin=5mm}
    \begin{tabular}{|c|p{4.5cm}|}
        
        \hline
        \textbf{query} & hows weather today \\
        \hline
        \textbf{bot response} & I'm sorry, but I don't have information about the weather. How can I assist you with the credit card? \\
        \hline
        \textbf{gpt4 eval} & Good \\
        \hline
        \textbf{gpt4 reason} & The answer admits that it does not have information about the weather and asks how it can assist with the the credit card. The answer is relevant to the query and meets the answerability criteria.\\
        \hline
        
        \end{tabular}
    \vspace{2mm}
    \caption{GPT-4 evaluation for an OOD query}
    \label{tab:gpt1}
\end{table}



We evaluated the bot output on a chat session's sequence of 25 queries. Queries consisted of English and Hinglish domain queries, OOD queries (e.g., hows the weather), greeting/acknowledgment queries, etc. The GPT-4 evaluation provided 100\% accuracy for these answers. A manual verification of evaluations was done to re-confirm the results. The evaluation indicated that the FAQ bot is able to respond accurately to user queries.



\section{Optimizing RAG for FAQ chatbots}

For publicly available API-based paid LLMs (e.g., ChatGPT), the cost is calculated based on the number of input and output tokens. The larger the number of tokens, the higher the cost. Also, it was observed that the accuracy of LLMs degraded with more tokens since LLMs are tasked with the additional task of identifying relevant tokens from the entire context \cite{liu2023lost}. LLM cost can be significant if many users are interacting with the bot. In this section, we propose two approaches to optimize the cost of LLMs without trading off the accuracy.

\subsection{Reinforcement learning for optimizing RAG}

In RAG, we fetch a top-k FAQ context for each query, pass it to an LLM, and generate the answer. However, there are patterns/sequences of queries for which not fetching a FAQ context may still lead to a good bot answer. This can happen under the following scenarios.

\begin{itemize}\label{it:nf}

  \item \textbf{Follow up queries}: 
  
  For a follow-up query, if the relevant context was fetched with the previous query, it would already be present with LLM. Hence, not retrieving a FAQ context for a follow-up query would still lead to the same answer.
  
  \item \textbf{Queries referring to the same FAQ}: 
  
  FAQs are designed to answer queries regarding a particular topic, e.g., in our FAQs, all the merchant cashback-related queries and benefits-related queries get answered from a FAQ 'what are the card benefits'. For the sequence of such queries referring to the same FAQ, we can fetch FAQ context only for the first query, and all the subsequent queries can utilize this context. Hence, fetching FAQ context on every query is unnecessary under such a scenario.  


  
 \item \textbf{OOD queries}: 
 
LLM is prompted to answer OOD queries with a "do not know" answer. Hence, even if we fetch FAQ context for OOD queries, LLM is instructed to ignore it. Hence, fetching a context would not be needed for OOD queries as well.

\end{itemize}

As can be seen from the list above, there are many cases for which fetching FAQ context would not be needed to get the same answer as with the usual RAG pipeline and, therefore, there is a scope to optimize the tokens passed to the LLM to reduce the cost without trading off the accuracy.





We propose a Reinforcement Learning (RL) based approach to optimize the number of tokens passed to LLM. Figure \ref{fig:pg} shows the proposed approach for policy-based agent for optimizing RAG. 
Specifically, we train a policy network to maximize rewards from a GPT-4 evaluator. The policy network resides outside the RAG pipeline and tries to optimize it by tuning the policy. We assume that we do not have access to gradients of the retrieval model or LLM and treat these components as fixed. 

Input to the policy network is the State, which comprises previous queries, the corresponding actions (encoded as tokens), and the current query. Since our RAG pipeline uses the last two conversations as history, for the policy model as well, we maintain the previous two queries as context. The policy network can take two actions for the current query: [FETCH] and [NO\textunderscore FETCH]. As the name suggests, when a policy network chooses a [FETCH] action, a usual RAG pipeline would be executed. When a policy network chooses a [NO\textunderscore FETCH] action, FAQ context is not retrieved, and the current query is directly inputted to the LLM.
The LLM then generates the answer. The current query, answer, and conversation history is inputted to a GPT-4, which rates the bot answer as good or bad. 

Intuitively, if a policy model chooses a wrong action, in that case, GPT-4 should provide a rating as Bad, e.g., if a policy model chooses a [NO\textunderscore FETCH] action for the in-domain query without any earlier context. We observe that it is indeed the case, and Table \ref{tab:gpt3} shows an example of such a case.

\begin{table}[!h]
    \centering
    \captionsetup{justification=centering, margin=5mm}
    \begin{tabular}{|c|p{4.5cm}|}
        
        \hline
        \textbf{query} & how much cashback on swiggy \\
        \hline
        \textbf{policy action} & [NO\textunderscore FETCH] \\
        \hline
        \textbf{bot response} & You will get 1.5\% unlimited cashback on Swiggy with the credit card. \\
        \hline
        \textbf{gpt4 eval} & Bad \\
        \hline
        \textbf{gpt4 reason} & The given answer is incorrect. According to the FAQ context, the credit card offers 4\% unlimited cashback on preferred partners, including Swiggy. The answer provided 1.5\% cashback, which is not accurate.\\
        \hline
        
        \end{tabular}
    \vspace{2mm}
    \caption{GPT-4 evaluation for an in-correct policy action}
    \label{tab:gpt3}
\end{table}

\begin{table}[h]
    \centering
    \captionsetup{justification=centering, margin=5mm}
    \begin{tabular}{|c|c|c|}
        
        \hline
        \textbf{Action} & \textbf{GPT-4 eval} & \textbf{Reward}\\
        \hline
        FETCH & - & 0.1 \\
        \hline
        NO\_FETCH & Good & 2 \\
        \hline
        NO\_FETCH & Bad & -1 \\
        \hline
        
        \end{tabular}
    \vspace{2mm}
    \caption{Reward shaping: converting GPT-4 evaluations to numeric rewards.}
    \label{tab:rw}
\end{table}

The GPT-4 evaluation rating is converted to the numeric reward ($r$) value as depicted in Table \ref{tab:rw}.
An intuition behind this reward function is that when a policy model executes a [NO\textunderscore FETCH] action and GPT-4 rates the bot response as Good, it indicates that for the input query not fetching a context still leads to a good output, possibly because one of the reasons mentioned in the itemized list. Hence, to promote such actions, we give a high positive reward. If the policy model plays a [FETCH] action, we give a small positive reward. If GPT-4 rates the bot response as Bad for either of the actions, we provide a negative reward to demote such actions, e.g., the policy model gets a negative reward when it does not fetch context when required. 
To minimize the cost for GPT-4 evaluations, we only evaluate the cases where a policy model plays a [NO\textunderscore FETCH] action. This is because a RAG pipeline is expected to work well with [FETCH] action, and we want to understand when executing a [NO\textunderscore FETCH] action is favorable. If an [FETCH] action leads to a wrong answer, then something within a RAG pipeline (embedding model/prompt) needs to be updated, and a policy model that resides outside the RAG will be unable to correct it. If the policy model plays a [FETCH] action, we directly assign a small positive reward without GPT-4 evaluation. Intuitively, we want a policy model to choose a [FETCH] action unless it is highly confident that a [NO\textunderscore FETCH] action will provide a correct answer.

Table \ref{tab:sar} shows a sample sequence of state, action, and rewards for the chat session with two queries ['is there an annual fee?', 'can you reduce it?'].

\begin{table}[!h]
    \centering
    \captionsetup{justification=centering, margin=5mm}
    \begin{tabular}{|c|p{5cm}|}
        
        \hline
        \textbf{$s_0$} & [CLS] is there annual fee [SEP] \\
        \hline
        \textbf{$a_0$} & [FETCH] \\
        \hline
        \textbf{$r_0$} & 0.1 \\
        \hline
        \textbf{$s_1$} & [CLS] is there annual fee [SEP] [FETCH] can you reduce it [SEP] \\
        \hline
        \textbf{$a_1$} & [NO\textunderscore FETCH] \\
        \hline
        \textbf{$r_1$} & 2. \\
        \hline
                
        \end{tabular}
    \vspace{2mm}
    \caption{Sequence of (State, Action, Reward) for a sample chat session with two queries}
    \label{tab:sar}
\end{table}

We calculate the cumulative sum of rewards $G_t$ for the $t^{th}$ query in a training chat session using the Eq. \ref{eq: total_reward}. 
\begin{equation}\label{eq: total_reward}
G_t = \sum_{k = 0}^N \gamma^kr_{t + k + 1}
\end{equation}
where $\gamma$ is the \emph{discount factor} and $N$ indicates the number of queries in a chat session. 
We set the value of $\gamma$ to a small value (0.1) since we care for an immediate reward, i.e., obtaining good output for each query. 

We train the policy model on six chat sessions consisting of 168 queries. Additional chat sessions are created by randomly shuffling the queries within a chat session. For each query within a session, we sample a [FETCH] or [NO\textunderscore FETCH] action based on the initial policy. The sampled action is executed, and a bot response is generated. GPT-4 then evaluates the bot response and provides a rating as Good or Bad, which is then converted to a numeric reward. We repeat this process multiple times per chat session and create a dataset of 1733 (state, action, rating) tuples. 

As a policy model, we use an in-house pre-trained BERT model. The in-house BERT model has the same architecture as bert-base-uncased, with 12 layers and an encoding dimension 768. The model is trained using Masked language modeling (MLM) and Next sentence prediction (NSP) on in-domain text such as product descriptions, user reviews, etc. We add a linear layer with Softmax activation on the last layer's embedding of the [CLS] token to map the State representation onto a 2-dimensional action space. We add two new tokens to the vocab representing policy actions, [FETCH] and [NO\textunderscore FETCH], and randomly initialize their embedding. The policy network is trained with the policy gradient loss and entropy regularization loss shown in Eq. \ref{eq:pg}. Entropy regularization is shown to help exploration and even leads to better optimization \cite{ahmed2019understanding}.

\begin{equation}
l_t = -\log \pi_\theta (a_t|s_t) G_t - \lambda \hspace{0.1cm} H(\pi_\theta (a_t|s_t))
\label{eq:pg}
\end{equation}

where $H$ indicates entropy and $\lambda$ indicates the entropy loss weight. We set $\lambda$ to 0.1 for all our experiments. 
$s_t$ indicates the state at $t^{th}$ time step and $a_t$ indicates action taken at the $t^{th}$ time step.

        
        

We test the policy model on an extensive test chat session of 91 queries, which includes an interleaved sequence of in-domain queries (how to apply for card), greeting/acknowledgment queries (hi, ok, cool, thank u), and OOD queries. 
To judge the number of token savings obtained with the proposed optimizations, we compare the number of tokens passed to an LLM with a usual RAG pipeline to the settings where we introduce the optimizations for RAG. The number of tokens for LLM calls was calculated using the tiktoken library. With a usual RAG pipeline, we fetch a FAQ context for each query. 

With a policy-based approach, the policy model provides a probability prediction over actions for each query. We use the Monte Carlo dropout method to estimate uncertainty and average the probabilities over ten predictions. A [FETCH] or [NO\textunderscore FETCH] action is selected based on the averaged probability. 

\subsection{Similarity threshold on top-1 score (SimThr)}

As described earlier, the in-house embedding model trained with infoNCE loss provides accurate similarity scores for FAQs and discriminate scores w.r.t. OOD queries. An LLM call can be avoided if the top-1 similarity score is large. Hence, if the top-1 cosine score is above a pre-defined threshold (0.92) for a user query, we directly output a corresponding static FAQ answer. We execute a usual RAG pipeline if a similarity score is less than a threshold. Note that using such a threshold would not be reliable with the general-purpose public embedding model.


\subsection{Results}

Table \ref{tab:ts} shows the token saving under different experimental settings. The threshold-based approach (SimThr) provides \textasciitilde{12\%} saving on the number of tokens. 
Introducing an additional policy-based dynamic context selection provides \textasciitilde{31\%} saving in the number of tokens. 
For all the settings, we performed a manual quality evaluation of the bot responses. We observe a slightly lower accuracy for All fetch and All fetch + SimThr setting than the SimThr + policy setting. In particular, we notice that for All fetch and All fetch + SimThr setting, for one of the acknowledgment queries ('cool'), the bot hallucinates and incorrectly outputs an answer for an FAQ. For the same query and context, a policy-based setting correctly outputs the relevant message.  

Note that even with a limited training set available for policy training, we achieved significant savings on the number of tokens using the combination of SimThr and a policy-based approach. Out of the entire token saving, \textasciitilde{61\%} can be attributed to the policy model, and \textasciitilde{39\%} can be attributed to the SimThr approach.




        
                

\begin{table}[h]
    \centering
    \captionsetup{justification=centering, margin=5mm}
    \begin{tabular}{|c|c|c|c|}
        
        \hline
        \textbf{Setting} & \textbf{\# tokens} & \textbf{saving} & \textbf{Acc.}\\
        \hline
        All fetch & 167885 & - & 0.989\\
        \hline
        SimThr & 147147 & \textasciitilde{12\%} & 0.989 \\
        \hline
        SimThr + policy & 115259 & \textasciitilde{\textbf{31\%}} & \textbf{1.}\\
        \hline
                
        \end{tabular}
    \vspace{2mm}
    \caption{Comparison of token savings with different experimental settings}
    \label{tab:ts}
\end{table}

We also experimented with a different reward shaping, in which, for [NO\textunderscore FETCH] action, we give a positive reward on the same scale as the negative reward. We observed a token saving of \textasciitilde{30\%} for the same test set, less than the earlier policy. This indicates that the reward shaping affects the token saving.

Table \ref{tab:ss} shows the result of the policy model on the sequence of three queries. In our FAQ dataset, all merchant cashback-related queries get answered from the same FAQ 'what are the card benefits'. For the sequence of these cashback-related queries, the policy model correctly retrieves context only for the first query, does not retrieve context for subsequent queries, and still provides the accurate answer. This underlines the effectiveness of the proposed RL-based approach for FAQ chatbots with RAG.

\begin{table}[!h]
    \centering
    \captionsetup{justification=centering, margin=5mm}
    \begin{tabular}{|c|p{5cm}|}
        
        \hline
        \textbf{query} & what cashback on i will get on swiggy  \\
        \hline
        \textbf{policy action} & [FETCH] \\
        \hline
        \textbf{bot answer} & You will get 4\% unlimited cashback on Swiggy with the credit card. \\
        \hline
        \textbf{human eval} & Good \\
        \hline
        
        \hline
        \textbf{query} & what about pvr \\
        \hline
        \textbf{policy action} & [NO\textunderscore FETCH] \\
        \hline
        \textbf{bot answer} & You will get 4\% unlimited cashback on PVR with the credit card.  \\
        \hline
        \textbf{human eval} & Good \\
        \hline

        \hline
        \textbf{query} & how much on curefit \\
        \hline
        \textbf{policy action} & [NO\textunderscore FETCH] \\
        \hline
        \textbf{bot answer} & You will get 4\% unlimited cashback on Curefit with the credit card.  \\
        \hline
        \textbf{human eval} & Good \\
        \hline
                
        \end{tabular}
    \vspace{2mm}
    \caption{Policy model predictions on a sequence of queries referring to the same FAQ}
    \label{tab:ss}
\end{table}

        
                

\subsection{Comparison with gpt-2 as policy model}

We compare the results obtained with the in-house BERT policy model with the publicly available gpt-2 model as a policy model. For gpt-2 model, apart from [FETCH] and [NO\textunderscore FETCH] tokens, we also add [CLS] and [SEP] tokens to the vocab and randomly initialize their embeddings. For gpt-2, we use the embedding from the last layer's [SEP] token and add a linear layer on top of it. All the remaining experimental settings are the same as the in-house BERT experiment. Table \ref{tab:gpf} shows the comparison result. Using the gpt-2 model as the policy provides \textasciitilde{25\%}  of token saving with the same accuracy. This indicates that the in-house BERT model performs better with the limited training set due to domain pre-training.

        
                       

\begin{table}[h]
    \centering
    \captionsetup{justification=centering, margin=5mm}
    \begin{tabular}{|c|c|c|c|}
        
        \hline
        \textbf{Setting} & \textbf{\# tokens} & \textbf{saving} & \textbf{Acc.}\\
        \hline
        SimThr + policy (gpt-2) & 124924 & \textasciitilde{25\%} & 1.\\
        \hline
        SimThr + policy (BERT) & 115259 &  \textasciitilde{31\%} & 1. \\
        \hline
                       
        \end{tabular}
    \vspace{2mm}
    \caption{Comparison of token saving with a open gpt-2 as policy model}
    \label{tab:gpf}
\end{table}

\section{Conclusion}

In this paper, we described a RAG-based chatbot for answering credit card-related queries using the domain FAQ dataset. We generated the question paraphrase data using the public LLM model and trained an in-house embedding model for retrieval using the infoNCE loss. An in-house model was shown to perform significantly better than a general-purpose public model in terms of ranking accuracy and OOD query detection. Further, we noticed that for specific patterns/sequences of queries, it is not required to fetch the FAQ context to get a good answer. Assuming a fixed retrieval model and LLM, we optimize the number of tokens passed to an LLM using Reinforcement Learning. We trained a policy model that resides external to the RAG and decides whether to retrieve a context. We used a GPT-4 as the reward model. The RL-based optimization combined with the similarity threshold led to significant token savings while slightly improving the accuracy. The proposed policy-based approach is generic and can be used with any existing RAG pipeline.

\bibliography{aaai-rl}

\end{document}